\documentclass{llncs}

\usepackage{geometry}
\usepackage{graphicx} 
\usepackage{csquotes}
\usepackage{enumitem}
\usepackage{multicol} 

\usepackage{mathtools}
\usepackage{fancyhdr}

\usepackage{hyperref}
\begin{document} 

\title{\texorpdfstring{Auto-survey Challenge: \\Advancing the Frontiers of Automated Literature Review\thanks{Both authors contributed equally. The authors are in alphabetical order of last name.}}{Auto-survey Challenge}}
\author{Thanh Gia Hieu Khuong, Benedictus Kent Rachmat}        
\institute{INRIA}

\maketitle

\begin{abstract} 
We present a novel platform for evaluating the capability of Large Language Models (LLMs) to autonomously compose and critique survey papers spanning a vast array of disciplines including sciences, humanities, education, and law. Within this framework, AI systems undertake a simulated peer-review mechanism akin to traditional scholarly journals, with human organizers serving in an editorial oversight capacity. Within this framework, we organized a competition for the AutoML conference 2023. Entrants are tasked with presenting stand-alone models adept at authoring articles from designated prompts and subsequently appraising them. Assessment criteria include clarity, reference appropriateness, accountability, and the substantive value of the content. This paper presents the design of the competition, including the implementation baseline submissions and methods of evaluation. 
\end{abstract} 

\vspace{-1cm}
\section{Introduction and motivation}

The rapid advancement of Large Language Models (LLMs) like GPT-3\cite{brown2020language} and Bard over recent years has resulted in machines capable of generating coherent, long-form text, prompting a paradigm shift in how we approach and leverage Artificial Intelligence (AI). This unprecedented level of proficiency\cite{cortes2021inconsistency} has opened up new avenues in various fields, including in academia. As LLMs get closer to human-level text generation, we are encouraged to explore their ability to autonomously produce and assess academic content. This prompted us to organize a competition, which has been selected as part of the official challenges of the AutoML conference 2023 \footnote{\url{http://auto-survey.chalearn.org/}}.

\vspace{-0.5cm}
\section{Challenge design}
The challenge proposes two tasks: AI-Author and AI-Reviewer. 

For the {\bf AI-Author task}, participants create topic-specific survey papers based on prompts (limited to 2000 words, including references).  
A typical prompt would be: ``Write a systematic survey or overview about the incorporation of writing assignments within the computer science curriculum''.
To generate prompts, we ``reverse engineered'' Semantics Scholar survey papers, by asking a language model to generate a prompt that would lead to generating such a survey paper. This lead to 80 prompts spanning a wide diversity of domains in sciences and humanities.

We implemented a baseline author based on ChatGPT~\cite{brown2020language} version GPT-3.5.
To evaluate the submissions of the AI-Author task, we implemented our own {\bf AI-Referee-Reviewer} (which also serves as a baseline for the AI-Reviewer task) using {\it ad hoc} publicly available software (e.g. \cite{vainshtein2019assessing,lees2022new} for clarity and responsibility). 

For the {\bf AI-Reviewer task}, participants' AI systems evaluate survey papers using predefined criteria, assigning review scores and justifications. 
 
To evaluate the reviews produced by the participants' code, we implemented a meta-reviewer, which criticizes the reviews.

The challenge has 3 phases: feed-back phase (to seek participant feed-back on the protocol), development phase (participants submit code to a challenge platform and are automatically evaluated), and final test phase (evaluation by a human jury of the final code submission on new prompts). We describe, in the next section, the methods implemented to perform automated evaluation in the development phase.

\vspace{-0.5cm}  
\section{Evaluation methods}\label{sec:eval}
The evaluation framework for survey papers is based on five key metrics: Relevance, Contribution, Soundness, Clarity, and Responsibility. ``Relevance'' checks whether the contents is consistent with the prompt. ``Contribution'' evaluates the comprehensive coverage of the survey. ``Soundness'' focuses on factual accuracy backed by authoritative references. ``Clarity'' evaluates readability through language use, paper structure, and clear communication of concepts. ``Responsibility'' checks for ethical considerations and adherence to moral values.

To standardize evaluations, reference ``good" and ``bad" versions of papers for each prompt were artificially generated with a language model, by paraphrasing the original human paper from which the prompt was derived.

\begin{figure}
    \centering
    \includegraphics[width=0.8\linewidth]{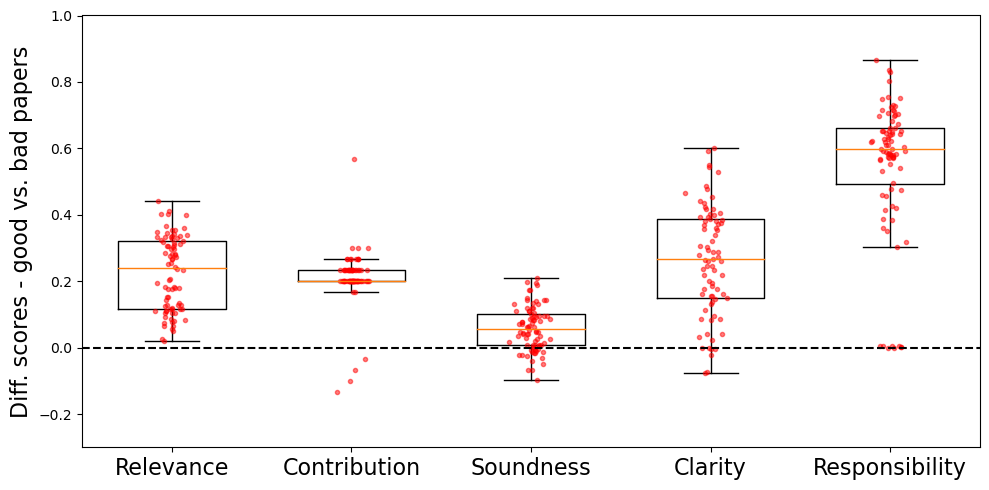}  
    \caption{``Contrastive evaluation'' of the AI-Referee-Reviewer using ``good" and ``bad" versions of survey papers}
    \label{fig:paired-t-test}
\end{figure}

To evaluate the AI-Reviewers (including our own AI-Referee-Reviewer)

we compare the reviews of the ``good" and ``bad" versions of survey papers we generated, for the various review criteria. 
Fig. \ref{fig:paired-t-test} represents the distribution of differences in review scores for pairs of good and bad papers for various criteria, which we call ``contrastive evaluation''.
Each red dot represents a pair of good and bad paper. The horizontal displacement for each box is here just to help for visualization and is not meaningful.  The distribution is also visualized with the box-whisker convention. The red line is the median of the distribution, the upper and lower ends of the boxes represent the quartiles, and the whiskers the 10 and 90th percentiles. 
Larger values indicate that the AI-Referee-Reviewer did a good job because it gave higher scores to good than to bad papers. Relevance, Clarity, and Responsibility are the criteria that are relatively easy to evaluate with our AI-Referee-Reviewer. For Relevance, we are please to see that the semantic similarity between prompt and text of the survey paper is easy to detect using Sentence Transformer~\cite{reimers2019sentence}. For Clarity and Responsibility, this reveals the maturity of the field in these areas, since we rely on state-of-the-art features and software \cite{lees2022new}. Contribution and Soundness are the two criteria which will require most future work. 

\vspace{-0.5cm}
\section{Baseline Results}

{\bf AI-Author Task Baselines:} 

\begin{figure}
    \centering
    \includegraphics[width=\linewidth]{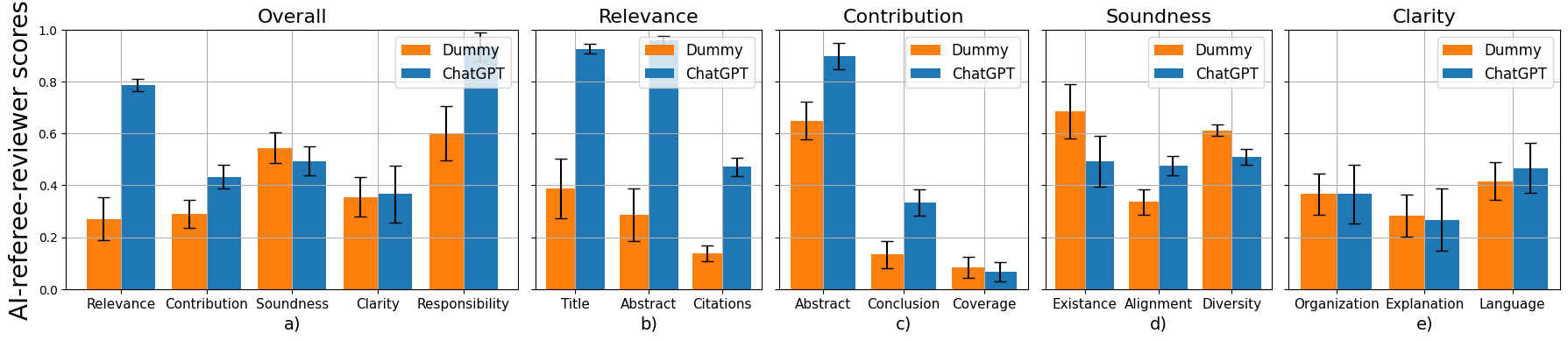}  
    \caption{Baseline AI-Authors, evaluated by the AI-Referee-Reviewer}
    \label{fig:chatgpt1}
\end{figure} 
\vspace{-0.5cm}
In this section, we evaluate our AI-Author baseline submission, built with ChatGPT, using our AI-Referee-Reviewer, and compare it to a ``Dummy'' baseline.
In Fig. \ref{fig:chatgpt1} we compare the various criteria for two types of papers: {\bf Dummy Baseline} (in ORANGE) is a baseline that returns random human papers regardless of prompts, hence irrelevant, but otherwise good. It achieved a review score of 0.15 (±0.07). {\bf ChatGPT Baseline} (in BLUE) is a baseline that uses ChatGPT to generate responses and achieved a review score of 0.50 (±0.06). This result showcases large language models' ability to create relevant and coherent responses.

In Fig. \ref{fig:chatgpt1}a, we show the normalized score values for the various review criteria, and criteria are broken up into sub-criteria (except for responsibility that has only one sub-criterion).  
If we look at the relevance sub-criteria in Fig. \ref{fig:chatgpt1}b, we see that indeed neither the title and abstract nor the citations are relevant to the prompt for the dummy papers but they are very relevant for the ChatGPT papers, hence the big gap. ChatGPT also dominates on Responsibility since the model is designed to be respectful by default. Contribution obtains interesting results in Fig. \ref{fig:chatgpt1}c. The language model, which is judging itself, is particularly happy about the way the abstract summarizes the text. The coverage, which is the alignment of the citations with those of the original human paper is bad for both types of papers. This is something on which we need to put more effort. Soundness, the quality of citations supporting a paper, is slightly higher in human-generated papers, as illustrated in Figure \ref{fig:chatgpt1}d. Finally, as expected, the gap for clarity in Fig. \ref{fig:chatgpt1}e is not large, since both are written in good English.

\begin{figure}
    \centering
    \hspace*{0.6in}
    \includegraphics[width=0.8\linewidth]{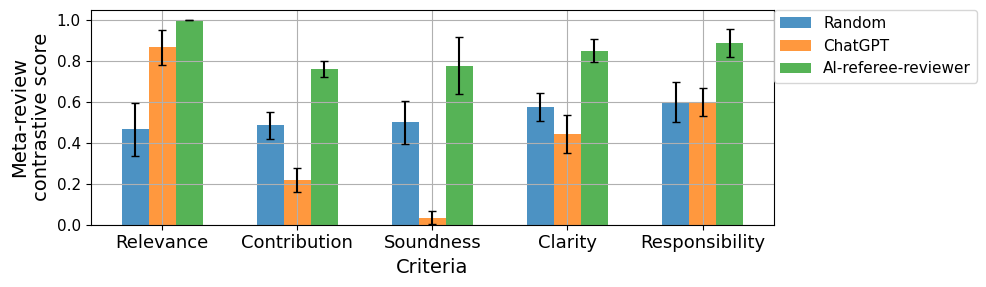}   
    \caption{Baseline AI-reviewers, evaluated by our meta-reviewer. }
    \label{fig:reviewer}
\end{figure} 
\vspace{-0.5cm}
{\bf AI-Reviewer Task Baselines: }

In this section, we assess the performance of various AI-reviewer baselines, using our ``contrastive evaluation'' approach.
In Fig. \ref{fig:reviewer}, we show the evaluation by our ``meta-reviewer'' for the various review criteria.
We plot the ``meta-review contrastive score'', which is the fraction of correct classification of ``good'' and ``bad'' survey papers. This is related to the contrastive evaluation (Section \ref{sec:eval}) conducted for the AI-Referee-Reviewer in Fig. \ref{fig:paired-t-test}. Indeed, for the AI-Referee-Reviewer, the ``meta-review contrastive score'' would correspond to the fraction of red points above the dashed line. Note that there are statistical fluctuations for each experimental run.

We analyze 270 papers in total. The error bars indicate the standard error calculated across 15 subsets of good/bad papers, each containing 18 good/bad versions generated from identical prompts.  

{\bf Random:} This baseline generated random reviews (random review criteron scores and random review text), resulting in a meta-review contrastive score always around 0.5, which is expected. {\bf ChatGPT:} All reviewer scores of this baseline were generated using ChatGPT. We notice that Relevance is rather well assessed by this reviewer. However, for all the other criteria, this reviewer is too lenient for bad papers, resulting in poor meta-review contrastive scores.
ChatGPT's weakest performance area is in terms of Soundness. When tasked with evaluating whether an answer provides precise information backed by citations from reliable and authoritative sources, it demonstrates similar or greater satisfaction with the fictitious references it creates for ``bad" papers, rather than with the genuine citations from the ``good'' papers.
{\bf AI-Referee-Reviewer:} This is our own implementation of a reviewer, also used to evaluate AI-authors in the challenge. It achieves the best performance of all our baselines, and is a target to be beaten by the challenge participants.
We obtain partcularly good results for the Relevance criterion, which we obtained using 
Sentence Transformers embeddings\cite{reimers2019sentence}. 
The other criteria need further improvements. 

\vspace{-0.5cm}
\section{Conclusion}
\vspace{-0.25cm}
Our baseline submission based on ChatGPT demonstrates the feasibility of the tasks we propose in this challenge. However, there is considerable room for improvement.
As of the submission of this paper, the competition has started and we already have a few submissions. We implemented a detailed tutorial with Colab notebooks and hope that this will encourage participants to enter this difficult challenge. We are open to organizing a tutorial and/or a hackathon at the JDSE, so students can get exposed to this exciting field.

\vspace{-0.5cm}
\subsection*{Acknowledgements}

This work was performed as part of an internship funded by INRIA, under the supervision of Isabelle Guyon. Support by ANR Chair of Artificial Intelligence HUMANIA ANR-19-CHIA-0022, TAILOR EU Horizon 2020 grant 952215, and Google are also gratefully acknowledged.

\iffalse
\textbf{AI-Author Task}: ChatGPT's responses are relevant and coherent, yet improvements are needed in addressing prompts comprehensively and providing high-quality citations.

\textbf{AI-Reviewer Task}: While models offer coherent feedback, generating precise, accurate, and comprehensive numeric evaluations remains challenging, partly due to the overly optimistic nature of the model.

The competition aims to bridge these gaps by challenging participants to create AI systems that write and review academic papers. The dual-task and peer-review structure of the competition serves as a comprehensive testbed for exploring AI's potential in academic content generation and critique.

The competition's future trajectory is anticipated to foster advanced AI models and systems capable of producing high-quality academic papers and insightful, accurate academic reviews. This progress will significantly impact academic research, artificial intelligence, and natural language processing, ushering in a new era of AI-assisted academic discourse. Through innovative approaches and solutions, the competition seeks to push AI's boundaries in academic writing and reviewing, creating a lasting influence on the scientific community.
\fi

\vspace{-0.5cm}
\bibliography{mylib}
\bibliographystyle{IEEEtran}

\end{document}